\documentclass[letterpaper, 10 pt, conference]{ieeeconf}  

\IEEEoverridecommandlockouts                              

\usepackage{mathtools}
\usepackage{amsmath, amssymb, dsfont}
\usepackage{bm}

\usepackage{graphics}
\usepackage{subcaption}
\usepackage{cite}
\usepackage{url}
\title{\LARGE \bf
Fingertip Contact Force Direction Control using Tactile Feedback
}

\author{Dounia Kitouni$^{1}$, Elie Chelly, Mahdi Khoramshahi and Veronique Perdereau
\thanks{*This work was partially supported by CORSMAL (CHIST-ERA program) under grant EP/S031715/1}
\thanks{All authors are with Institut des sytemes intelligents et de la robotique, Sorbonne University,
        4 Place Jussieu,75005, Paris, France}
\thanks{$^{1}$ The corresponding author 
         {\tt\small kitouni@isir.upmc.fr}}
}
\begin{document}

\maketitle
\thispagestyle{empty}
\pagestyle{empty}

\begin{abstract}

The human hand is an immensely sophisticated tool adept at manipulating and grasping objects of unknown characteristics. Its capability lies in perceiving interaction dynamics through touch and adjusting contact force direction and magnitude to ensure successful manipulation. Despite advancements in control algorithms, sensing technologies, compliance integration, and ongoing research, precise finger force control for dexterous manipulation using tactile sensing remains relatively unexplored.
In this work, we explore the challenges related to individual finger contact force control and propose a method for directing such forces perceived through tactile sensing. The proposed method is evaluated using an Allegro hand with Xela tactile sensors. Results are presented and discussed, alongside consideration for potential future improvements.

\end{abstract}


\section{INTRODUCTION}

Contact force control enhances the robot's ability to handle physical contact and improves the overall manipulation task.
Contact with the environment in robotic manipulations was long considered tedious and complex to manage \cite{suomalainen2022survey}.
This is mainly due to the changing contact dynamics encountered when transitioning from having no contact with the object to contact and vice versa, the state of contact itself, i.e., contact with or without slippage and when lifting an object. 
Another component is environmental uncertainties arising from imprecise or unknown object characteristics such as shape, weight, and friction. 
Moreover, contact models used to model the interaction between the robot and the object are complex and nonlinear. When implementing force control strategies, these contact models are often simplified to ease calculation, adding an extra layer of environmental uncertainty.

Contact force control witnessed significant advancement led by improvements in sensing, adaptive control algorithms, robot compliance, either mechanical or virtual, and ongoing research efforts.
Enabling robots to perform tasks requiring force modulation and adjustments such as peg-in-hole assembly, polishing, wiping, and opening doors \cite{A.Kramberger, Leidner2019wip,deng2016wavelet}.
However, while various strategies and techniques were developed to address the challenges of contact force control using tactile feedback, individual finger contact force control for robotic manipulation is still relatively unexplored \cite{deng2020grasping}.\\

Robots with individual finger control can perform a broader range of manipulation tasks that involve intricate object manipulation, such as turning a key, assembling small components, or handling delicate objects.
Furthermore, controlling each finger individually enables better force regulation and compliance, which is crucial when manipulating deformable objects or when the force applied to each contact point needs to be finely tuned.
Individual finger control provides the flexibility required to perform precise and delicate movements.

Enabling individual finger control in robotic hands brings them closer to human dexterity.
Studies in neuroscience showed that subtle finger movements are introduced during human manipulation that changes the direction of the forces applied to the manipulated object or the contact point\cite{johansson2009coding}. These changes modify the overall distribution of the forces applied to the object and enhance the grasp \cite{Goodwin_1998}.
Achieving precise control over the individual fingers of a robotic hand is a complex task in terms of both hardware and software. Developing sophisticated robotic hands capable of such control involves intricate mechanisms, sensors, and actuators\cite{billard2019trends}.

This study focuses on controlling the direction of contact forces for individual fingers using tactile feedback. The principal contributions of this work are as follows: 
\begin{enumerate}
    \item Proposing a method to control the direction of contact forces for individual fingers using tactile sensing.
    \item Experimentally testing and evaluating the proposed method and exploring its strengths and limitations.
    \item Providing insights on how individual contact force control can be improved.
\end{enumerate}
This article is structured as follows: First, we provide an overview of studies relevant to this field. Next, we describe the proposed method for controlling the direction of contact force using tactile feedback. We then present the protocol used to implement and test the proposed method, using an Allegro hand equipped with Uskin tactile sensors. Afterward, results from the tests are shown and discussed. Finally, we offer insights to refine further and advance this work.

\section{BACKGROUND}

The manipulation task is characterized by a dynamic interaction between the robot and the environment, in which the use of a pure motion control strategy for controlling interaction is prone to failure \cite{khatib2016springer}.
Successful execution of interaction tasks requires a focus on force control, typically categorized into direct and indirect force control methods.
Indirect force control relies on motion control to achieve force control without explicitly using force feedback.
On the other hand, direct force control offers the possibility of controlling contact force to a desired value thanks to the force feedback in the loop\cite{villani2016force}.
Indirect force control includes compliance and impedance control \cite{hogan1984impedance}, which uses mechanical stiffness or adjustable parameters impedance to relate position error to contact force.
Direct force control methods, like hybrid position/force control \cite{raibert1981hybrid,perdereau1993new}, require force/torque feedback and a detailed environment description. They are employed to control position along the unconstrained task directions and force along the constrained task directions \cite{Sciciliano2000}.

Many works have been done on force control, making it a well-established subject, including \cite{Gao_2020,Stepputtis_2022,Pang_2022,Yang_2022,Katayama_2022,Nakatsuru_2023}. However, if we take a closer look at these works, most of them are for industrial robot manipulators rather than for anthropomorphic robot hands. More attention should be paid to the force control of this kind of effectors, especially when we want them to perform dexterous manipulation \cite{nguyen2013fingertip}. 
Contact force was considered in methods for grasping force optimization \cite{liu2004real}, in which contact forces are obtained by projecting joint torques through a contact model such as contact with friction \cite{coulomb1809theorie}. Notably, these methods do not actively control contact forces but provide a combination of required forces to maintain a secure grip on an object, relying on contact models without real-time tactile force feedback \cite{Cloutier_2018}.

Various approaches have been employed to obtain information about physical interaction using tactile sensing.\cite{nguyen2013fingertip} proposed a method that models contact force and controls stiffness by estimating the contact location with tactile sensing. \cite{ramon2013finger} increased joint torques of manipulators when the pressure provided by tactile sensors decreased. \cite{Hang_2016} developed a method that used tactile sensing to control the impedance of the finger to adapt to environmental uncertainties.\cite{veiga2020grip} proposed a method that uses tactile sensing to detect slippage during grasp tasks and increases the amplitude of the normal forces in case of slips.
However, to the knowledge of the author, no previous work investigated the control of the direction of the contact force using tactile sensing feedback, which is an essential component since it allows the robot to adjust its finger to local friction conditions to prevent slips and to do fine manipulation tasks such as the control of the orientation of the object.

\section{PROPOSED METHOD}

In this section, we will present a method for controlling the direction of contact force. However, this can only be done when the contact between the robot's finger and the object is maintained and without slippage. In previous work, we proposed and validated a method for detecting individual fingertip slippage, which will first be presented shortly. Subsequently, the method for controlling the finger force direction will be detailed.

\subsection{Slippage Detection}\label{sec:contact_stab}
In our previous research \cite{kitouni2023model}, we addressed the problem of slippage detection at the finger level. 
we propose a classification method that distinguishes between stable contact, in which the contact between the fingers and the object is maintained, and unstable contact, which includes slippage and contact loss.
we hand-crafted features from tactile sensor feedback and formed a stable and unstable contact dataset. We use this data to train a classification model: Logistic regression. The model outputs a probability of the contact being stable $p$. The contact state is described by $y$:

\begin{equation}
    \left\{ \begin{array}{l}
         y = 1 \text{ if } p \geq 0.5 \\ 
         y = 0 \text{ if } p < 0.5 \end{array} \right. \label{eq_contact}
\end{equation}

$y=0$ denotes an unstable contact and $y=1$ a stable contact. When tested with previously unseen data, the classifier showed satisfactory performance with an accuracy of $95.5\%$.

\subsection{Contact Force Direction Control}

\begin{figure}[h]
    \centering
    \includegraphics[scale=0.4]{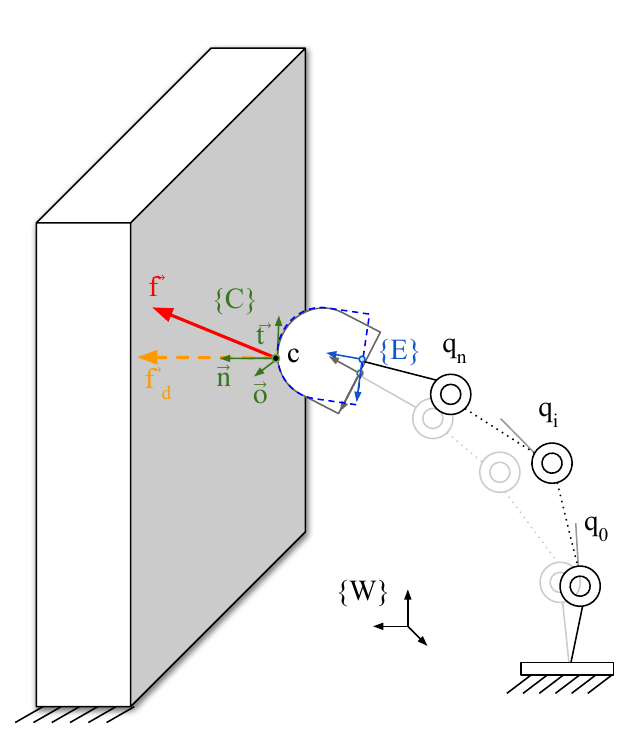}
    \caption{A robotic manipulator that is in contact with a rigid body}
    \label{fig:rigidbody}
\end{figure}

Consider a robotic finger in contact with a rigid body as in Fig.~\ref{fig:rigidbody}. The last segment of the finger will be referred to as the end-effector. 
Let $\{W\}$ represent the inertial frame fixed in the workspace and $\{E\}$ the fixed frame attached to the end-effector. 
Contact between the finger and the rigid body may occur at any location on the external surface of the end-effector and the object. 
$\{C\}$ is the contact frame with axes $ \{  \vec{o} , \vec{t}  , \vec{n} \}$. 
The unit vector $ \vec{n} $ is normal to the surface of the end-effector. The other two unit vectors $\vec{t} $ and $\vec{o}$ are orthogonal and lie in the contact tangent plane. 

$[p^T \phi_p^T]^T \in \mathds{R}^6$ represents the vector describing the position $p$ and the orientation $\phi_p$ of the end-effector frame  $\{ E \}$ ( end-effector pose ) relative to $\{ W \}$.
${[c^T,\phi_c^T]}^T \in \mathds{R}^6$ represents the vector describing the position $c$ and the orientation $\phi_c$ of the contact frame $\{ C \}$ relative to the end-effector frame $\{ E \}$. Lastly, $q=[q_1,..,q_i,..,q_m]^T \in \mathds{R}^m$ and $\tau \in \mathds{R}^m$ denote joint positions, and  joint torques, respectively.

Let $\prescript{\{W\}}{}{\mathbf{R}}_{\{E\}} \in \mathds{R}^{3\times3}$ be the rotation matrices that map the axis of the end-effector to the inertial frame $\{W\}$, and let $\prescript{\{E\}}{}{\mathbf{R}}_{\{C\}} \in \mathds{R}^{3\times3}$ be the rotation matrices that map the axis of the contact frame to the end-effector frame. 
Let $\prescript{\{C\}}{}{f}=[f_x,f_y,f_z]^T \in \mathds{R}^3$ be the force feedback measured by the tactile sensors in the contact frame $\{C \}$. it represents the forces arising from contact points between the robot and its environment.\\

\begin{equation}
       \prescript{\{W\}}{}{f}= \prescript{\{W\}}{}{\mathbf{R}}_{\{E\}}  \prescript{\{E\}}{}{\mathbf{R}}_{\{C\}} \prescript{\{C\}}{}{f}           
\end{equation}

We want to regulate the direction of the force feedback $\prescript{\{W\}}{}{f}$ and send it in a desired direction $\prescript{\{W\}}{}{f_d}$. We compute $R_\theta \in \mathds{R}^{3\times3}$ the rotation that maps $\prescript{\{W\}}{}{f}$  to the desired force feedback $\prescript{\{W\}}{}{f_d}$, following the the Rodriguez's formula\cite{Rodrigues} (more details on computation of $R_\theta$ are given in annex).

\begin{equation}
 \mathbf{R}_\theta= I^{3\times3} +s_\theta S(w) + (1- c_\theta) S(w)^2
\end{equation}

\cite{caccavale1998resolved} state that the orientation error  $\prescript{\{W\}}{}{R_\theta}$ can be expressed in terms of an axis angle representation $o$ in the inertial frame $\{W\}$ as:

\begin{equation}\label{eq:delta0}
   \prescript{\{W\}}{}{o} = \theta r 
\end{equation}

Where $\theta \in \mathds{R}$ and $r\in \mathds{R}^3$ are the angle and the axis respectively obtained from $R_\theta$. $r$ is normalized to obtain a unit vector.

In order to align $f$ with $f_d$, we need to rotate the end effector around the contact point. For this end, we generate $F_{task}$ at the end effector. We propose $F_{task}$ to be proportional to the end-effector's orientation  error as:

\begin{equation}\label{eq:delta1}
  F_{task}= K_\theta \prescript{\{W\}}{}{\Delta \phi}  
\end{equation}

Where $K_\theta \in \mathds{R}^{3\times3}$ is a diagonal gain matrix and $\Delta \phi$ is the orientation error of the $\{E\}$ frame in $\{W\}$. The $\{E\}$ frame orientation error relative to $\{W\}$, can be expressed as the axis angle error as: 

\begin{equation}\label{eq:delta2}
  \prescript{\{W\}}{}{\Delta \phi} = \prescript{\{W\}}{}{o}
\end{equation}

respectively, it can be expressed in ${\{C\}}$ as :
\begin{equation}
  \prescript{\{C\}}{}{\Delta \phi} = (\prescript{\{W\}}{}{\mathbf{R}}_{\{E\}}  \prescript{\{E\}}{}{\mathbf{R}}_{\{C\}})^T\prescript{\{W\}}{}{\Delta \phi}
\end{equation}

This mapping is only valid if the coordinates of the contact $[c^T, \phi_c^T]^T$ relative to $\{E\}$ remain constant, thus maintaining a fixed $\prescript{\{E\}}{}{\mathbf{R}}_{\{C\}}$.On the other hand, the coordinates of end-effector $[p^T, \phi_p^T]^T$ relative to $\{W\}$ may change. 
This process ensures a pure rotation around the contact frame without introducing any translation.
It's imperative to emphasize that our objective is not to precisely control the end effector's pose. Instead, we aim for displacements of the end-effector's pose to result from the rotation around the contact frame. 

Considering the quasi-static assumption \cite{kao2016quasistatic}, we can then translate $F_{task}$ into joint torques $\tau_{task}$ as follows:

\begin{equation}\label{eq:quasi}
  \tau_{task}={J_\omega}^T F_{task}   
\end{equation}
By substituting Eq.~\ref{eq:delta0}, Eq.~\ref{eq:delta1} and Eq.~\ref{eq:delta2} in Eq.~\ref{eq:quasi}, we obtain:
\begin{equation}
    \tau_{task}={J_\omega}^T k_{\theta} \theta r 
\end{equation}

The following equation governs the finger's dynamics:
\begin{equation}\label{eq:motion}
M(q) \ddot{q} + C(q,\dot{q})\dot{q} + g(q) + J^T F_{ext} = \tau_{\text{motion}} + \tau_{\text{task}}
\end{equation}

Where $M(q)$ is the mass matrix, $C(q,\dot{q})$ is the Coriolis effect, $g(q)$ is the gravitational effect and $F_{ext}$ are the forces arising from contact with the object. $F_{ext}$ equal to zero in the absence of contact. The torque $\tau_{\text{motion}}$ is associated with the motion of the finger, while $\tau_{\text{task}}$ incorporates additional task-related torque. In scenarios involving slow motion, the Coriolis effect becomes negligible, simplifying Equation \ref{eq:motion} as follows:

\begin{equation}\label{eq:motion_simplified}
M(q) \ddot{q} + g(q) + J^T F_{ext} = \tau_{\text{motion}} + \tau_{\text{task}}
\end{equation}

Here, $\tau_{\text{motion}}$ serves to move the finger toward the object, and $\tau_{\text{task}}$ corrects the orientation of the force.

Before correcting the orientation of $f$, a stable contact with the object must be established. We use $y$, the output of the contact stability detection method from Sec.~\ref{sec:contact_stab} to switch between fingertip position control and interaction force direction control. A Proportional-Derivative (PD) controller with inertia and gravity compensation allows incremental adjustment of joint positions to close the finger and make contact with the object. Upon achieving stable contact and receiving contact force feedback, $\tau_{\text{task}}$ is introduced to control the force direction and compensate for gravity. $y$ is the variable representing the type of contact. $y=1$ if there's stable contact between the robot and the object, and zero otherwise. Then, we choose our control as follows:

\begin{align}
\tau_{motion} &= (1-y) M(q) [K_p e(t) + K_d \dot{e}(t)]+ g(q) \label{eq_motion} \\
\tau_{task} &= y J_{\omega}^T K_{\theta} \theta r
\end{align}

Here, $K_p$ and $K_d$ represent the proportional and derivative gains of the PD controller, respectively, with $e(t)$ denoting the joint position error. To maintain contact while adjusting the force direction, a small force in the normal direction of the contact $n$ is added, modifying our control as follows:

\begin{align}
&\tau_{task} = y  [ J_{\omega}^T K_{\theta} \theta r +  J_{\omega}^T K_s  ( \prescript{\{W\}}{}{\mathbf{R}}_{\{E\}}  \prescript{\{E\}}{}{\mathbf{R}}_{\{C\}})  \prescript{\{C\}}{}{n} ] \label{eq_taks}
\end{align}
$K_s \in \mathds{R}^{3\times3}$ is a diagonal gain matrix representing contact stiffness, whose values we empirically chose to apply small force in the normal direction $n$.

\section{Experimental Evaluation}

\begin{figure}[h]
  \centering
  \begin{subfigure}{0.49\linewidth}
    \centering
    \includegraphics[height=0.8\linewidth,trim={4cm 5cm 4cm 10cm},clip]{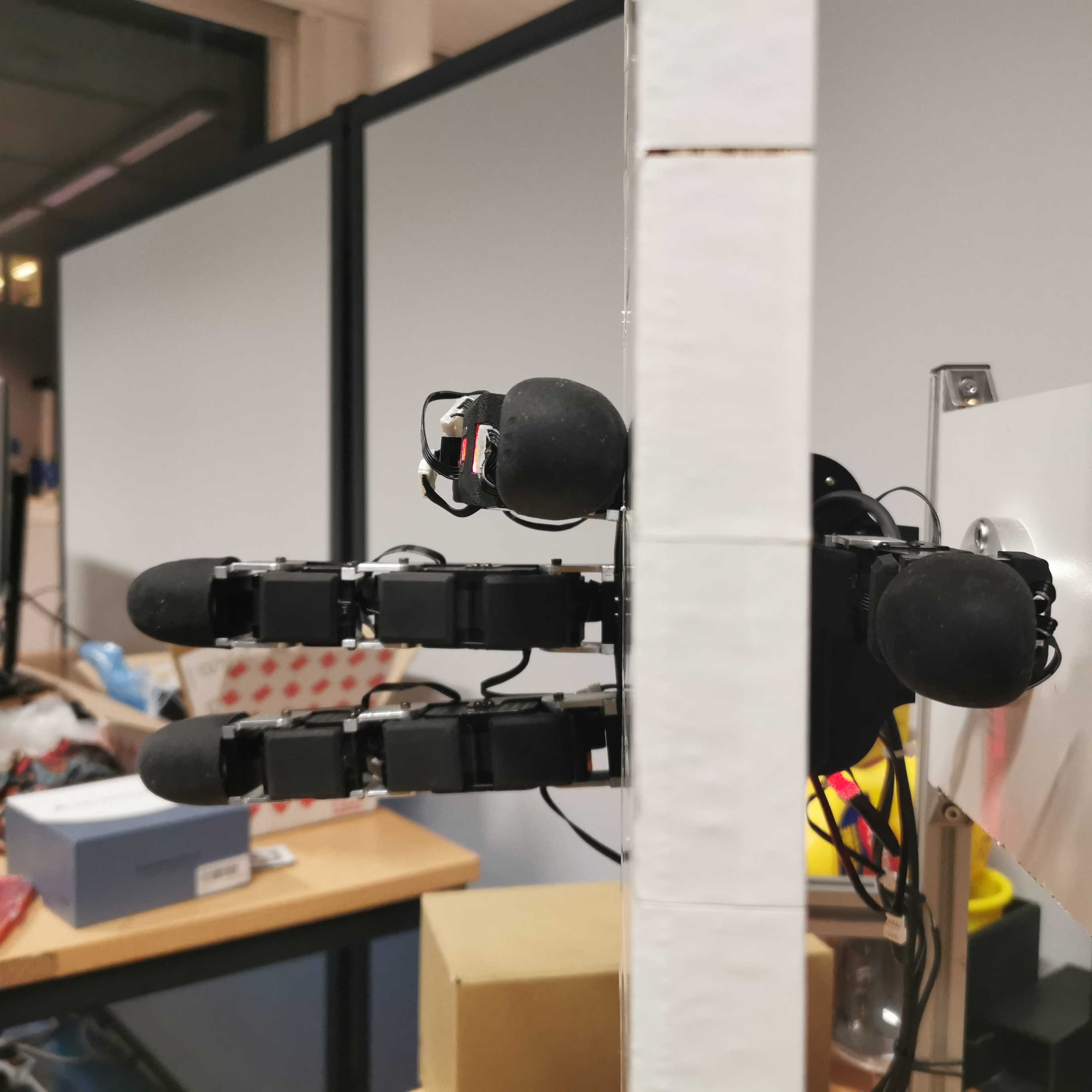}
  \caption*{}
  \end{subfigure}
  \hfill
    \begin{subfigure}{0.49\linewidth}
    \centering
    \includegraphics[height=0.8\linewidth,angle=180,trim={1cm 3cm 0.5cm 0},clip]{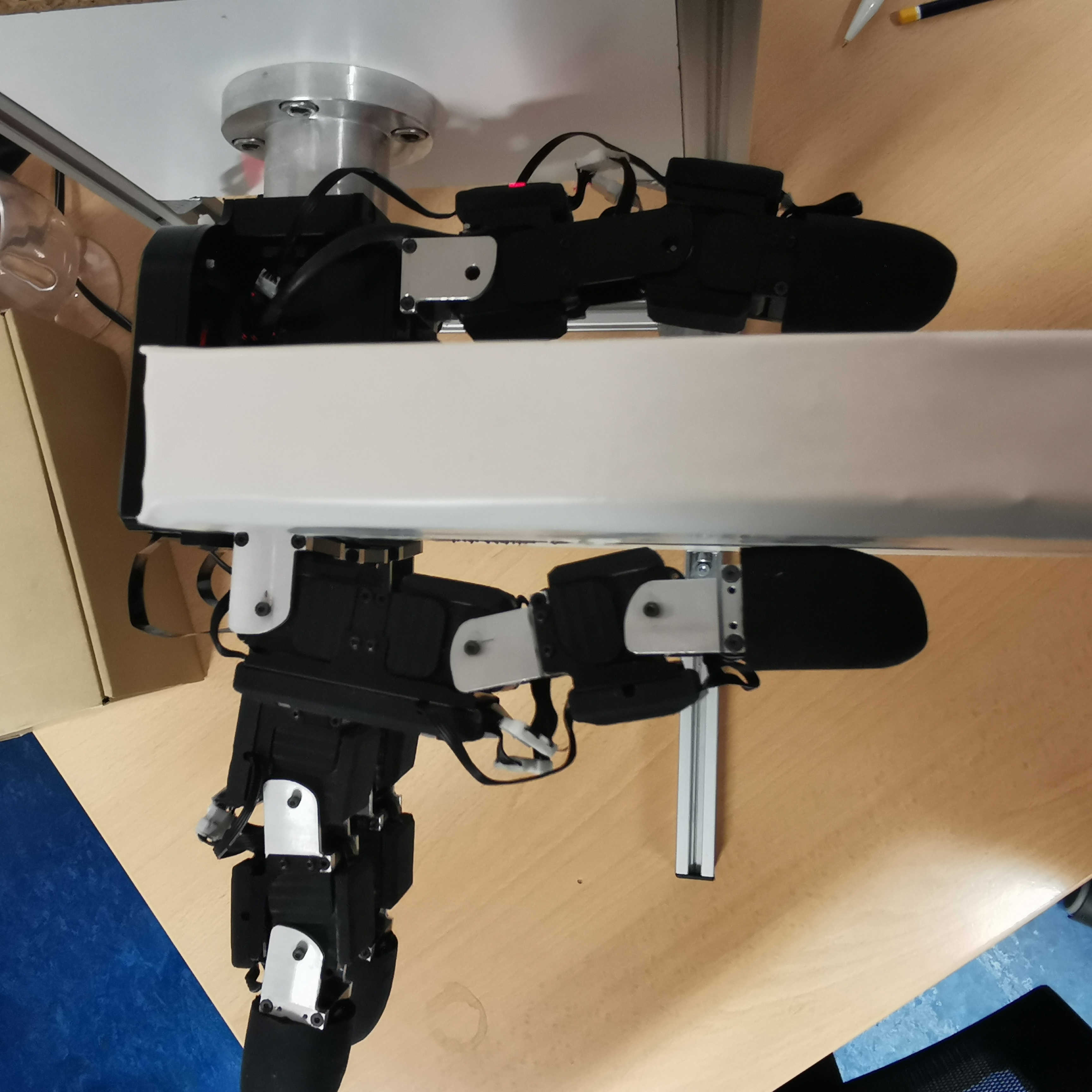}   \caption*{}
  \end{subfigure}
  \hfill
  \begin{subfigure}{0.49\linewidth}
    \centering
    \includegraphics[height=\linewidth,trim={0 3cm 0 2cm},clip]{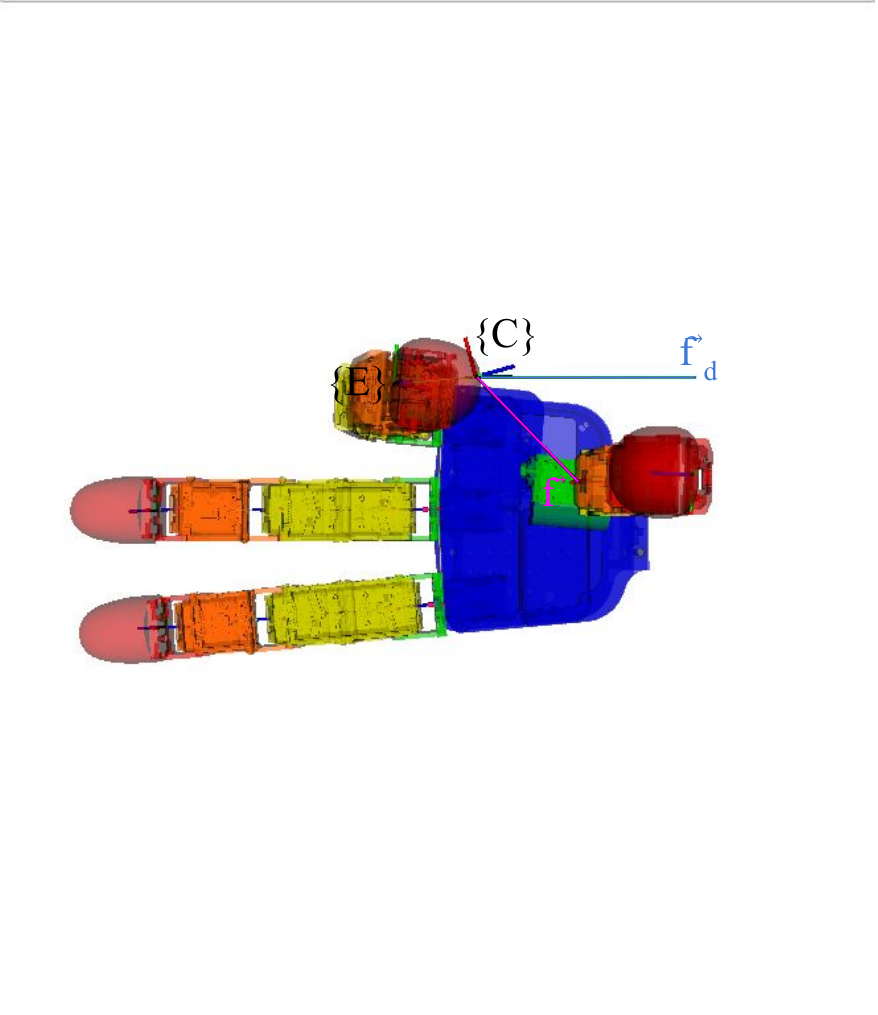}
   \caption{}
  \end{subfigure}
    \hfill
    \begin{subfigure}{0.49\linewidth}
    \centering
    \includegraphics[height=0.8\linewidth,trim={1cm 3cm 0.5cm 2cm},clip]{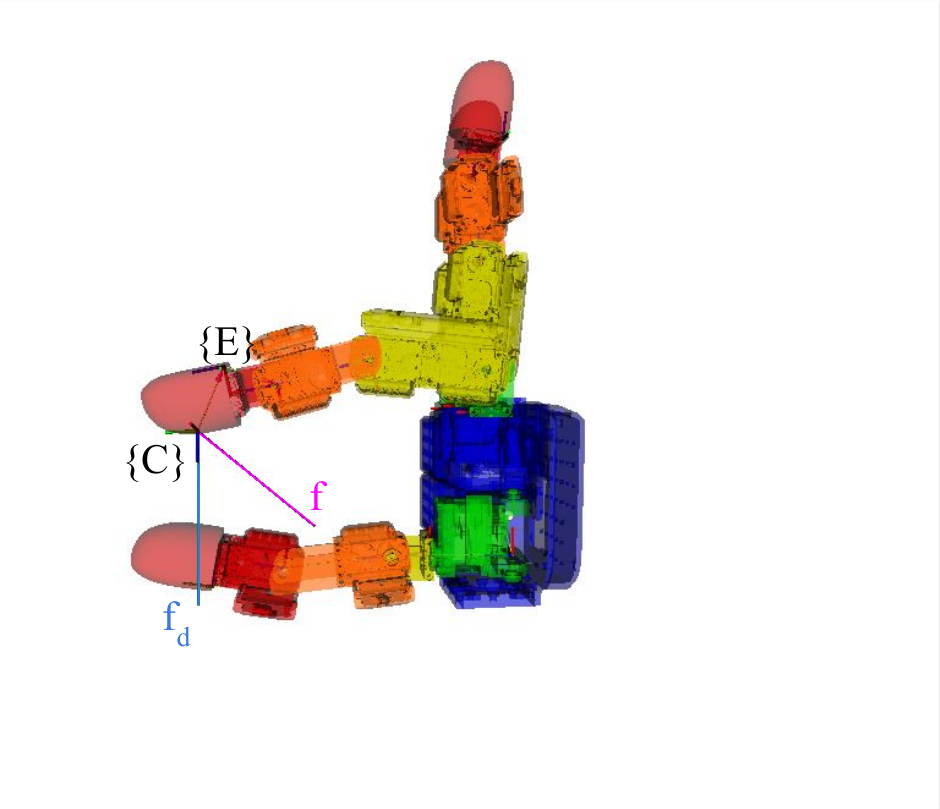}
   \caption{}
  \end{subfigure}
  \caption{The Allegro hand with tactile sensors and a 3D model view to show taxels positions.}
  \label{fig:allegro_hand_setup}
\end{figure}

We use the Index and the Thumb of an Allegro hand from Wonik Robotics \footnote{\url{http://wiki.wonikrobotics.com/}} to implement and test our method. It has 16 independent torque-controlled joints, 4 per finger. Each fingertip is equipped with uSkin soft sensors from Xela robotics \footnote{\url{https://www.xelarobotics.com/}} which are formed of 30 tri-axial taxels \cite{tomo2017covering}. Additionally, a rigid body is securely fixed within the hand's workspace see Fig~\ref{fig:allegro_hand_setup}. This configuration ensures a stable and controlled environment for testing the proposed control system. which allows for consistent and repeatable experiments, enabling us to evaluate the performance and proposed method precisely.

\subsection{ Contact Pose}
Each of the fingertips of the allegro hand is equipped with $n_{tx}=30$ taxels. The position and the orientation of each taxel is: 

\begin{equation}
 u_k=[ x_k\ y_k\ z_k\ \alpha_k\ \beta_k\ \gamma_k] ^T  \quad k=1,..,n_{tx}
\end{equation}

\begin{figure}[ht]
    \centering
    \includegraphics[scale=0.25,angle=90,trim={11cm 0 11cm 0},clip]{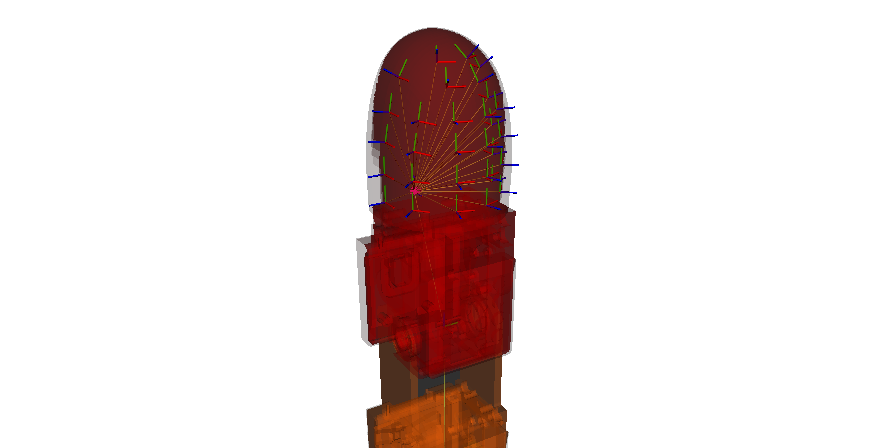}
    \caption{Taxel frames on a fingertip.}
    \label{fig:tx_frames}
\end{figure}

$x_k$, $y_k$, and $z_k$ the position of the origin of the taxes relative to the fingertip frame $\{E\}$ (Figure.~\ref{fig:tx_frames}). $\alpha_k$ $\beta_k$  and $\gamma_k$ represent the orientation of the contact frame relative to the fingertip frame $\{E\}$ in the roll pitch yaw representation.

Each taxel output $f_k=[f_{k,x}\ f_{k,y}\ f_{k,z}]^T$ is a measure of displacement in the taxel frame, which we will refer to as pseudo-force.

To obtain the position $ c=[ x_c\,y_c\, z_c ]^T$ and orientation of the contact frame $\phi_c=[\alpha_c\,\beta_c\,\gamma_c] ^T$ relative to $\{E\}$, we compute a weighted sum of the position of the taxels, as: 
\begin{equation}
    [c^T,\phi_c^T]^T=\frac{\sum_{i=1}^{n_{tx}} \Delta_k u_k }{\Delta}
\end{equation}

with:
\begin{equation}
    \left\{ \begin{array}{l}
         \Delta_k = || f_k || = \sqrt{f_{k,x}^2+f_{k,y}^2+f_{k,z}^2} \\ 
         \Delta = \frac{\sum_{k=1}^{n_{tx}} \Delta_k}{n_{tx}} \end{array} \right.
\end{equation}

\subsection{ Contact pseudo-force estimation }
\label{sec:contact_frc_estimation}
Now that we have the position $c$ and the orientation $\phi_c$ of the contact frame $\{C\}$, 
we can calculate the total pseudo-force $f \in \mathds{R}^3$ measured by the sensors, then project it to the contact frame as follows:

\begin{equation}
    f=R(\alpha_c,\beta_c,\gamma_c)^T \sum_{k=1}^{n_{tx}} R(\alpha_k,\beta_k,\gamma_k) f_k \label{eq_force}
\end{equation}

With $R$, a rotation matrix whose yaw, pitch, and roll angles are $\gamma_c$ , $\beta_c$ , $ \alpha_c$, respectively.

\subsection{Controller implementation}
The controller is implemented and tested for the thumb and the index fingers. The torque $\tau_{cmd} \in \mathds{R}^4$ applied is:
\begin{equation}
    \tau_{cmd} = \tau_{motion} + \tau_{task}
\end{equation}
With $\tau_{motion}$ described in Eq.~\ref{eq_motion} and $\tau_{task}$ described in Eq.~\ref{eq_taks}. The gain matrices $K_p, K_d, K_{\theta}$ and $K_s$ are tuned manually to reach the desired behavior. The gains that were settled for are different for each finger.

\section{Results and discussion}

\begin{figure}[h!]
    \centering
    \includegraphics[width=\linewidth]{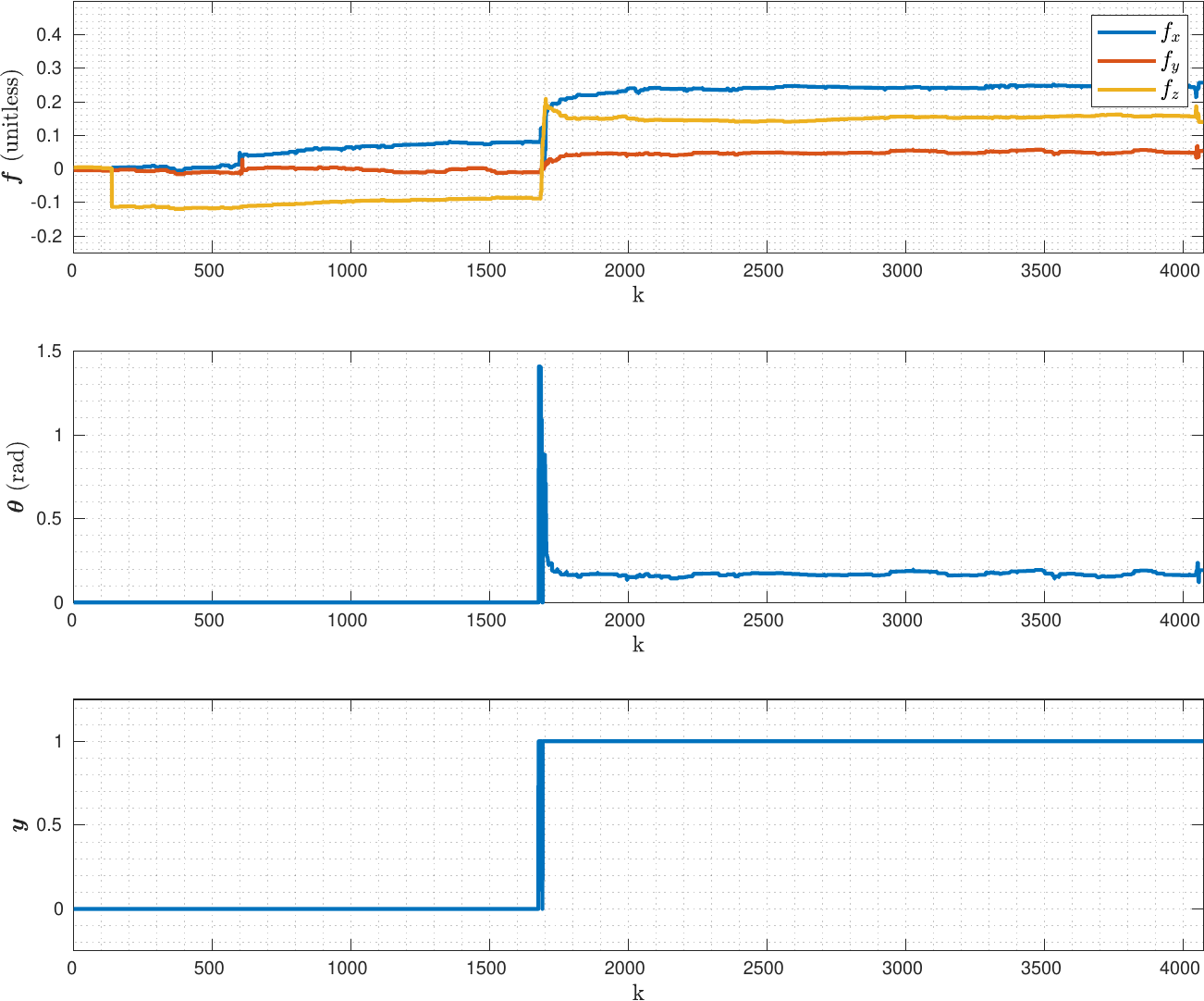}
    \caption{Results for contact force control for the Index}
    \label{fig:id_correction_1}
\end{figure}

\begin{figure}[h!]
    \centering
    \includegraphics[width=\linewidth]{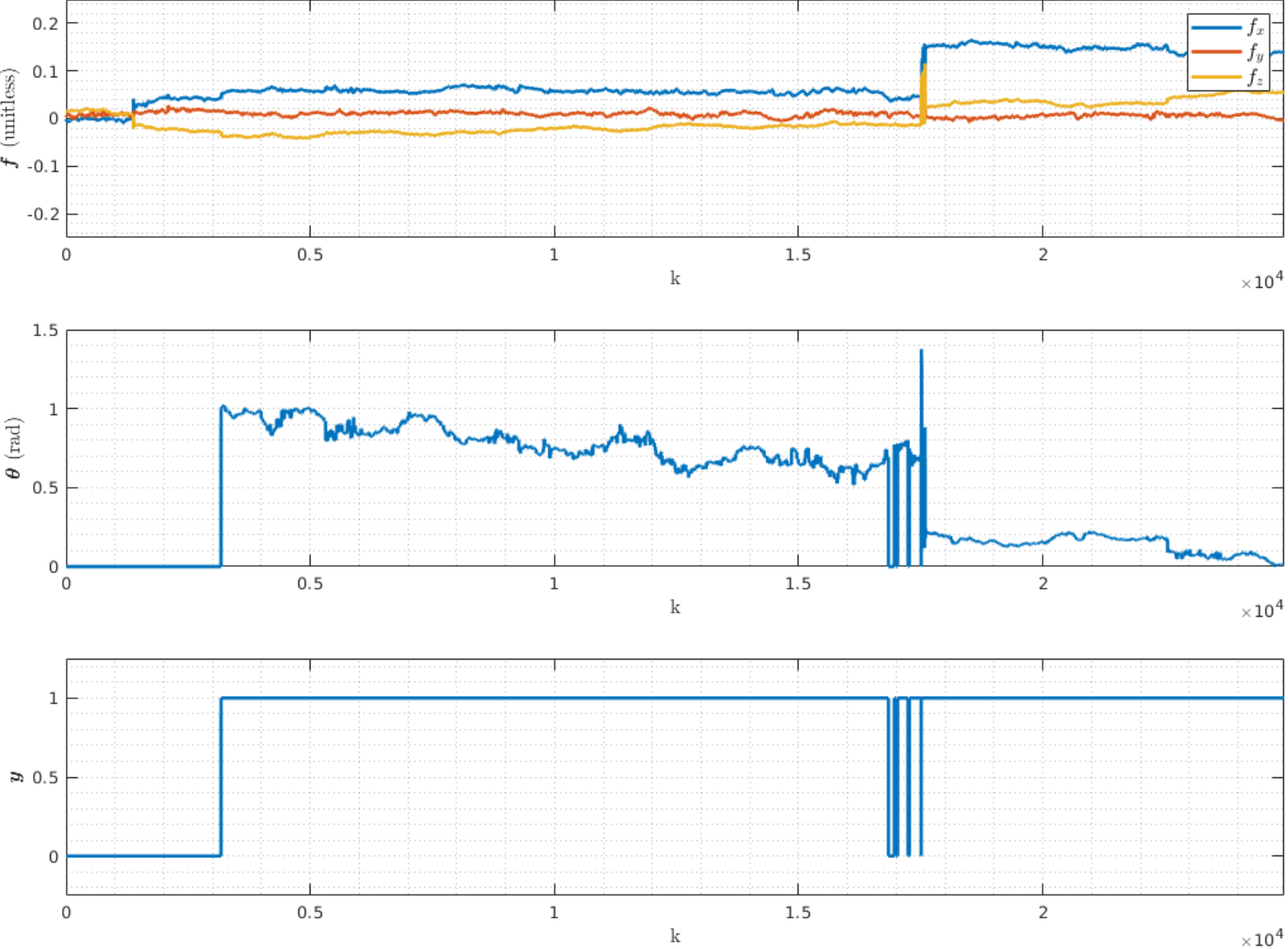}
    \caption{Results for contact force control for the Thumb}
\label{fig:md_correction_1}
\end{figure}

\begin{figure}[h!]
    \centering
    \includegraphics[width=\linewidth]{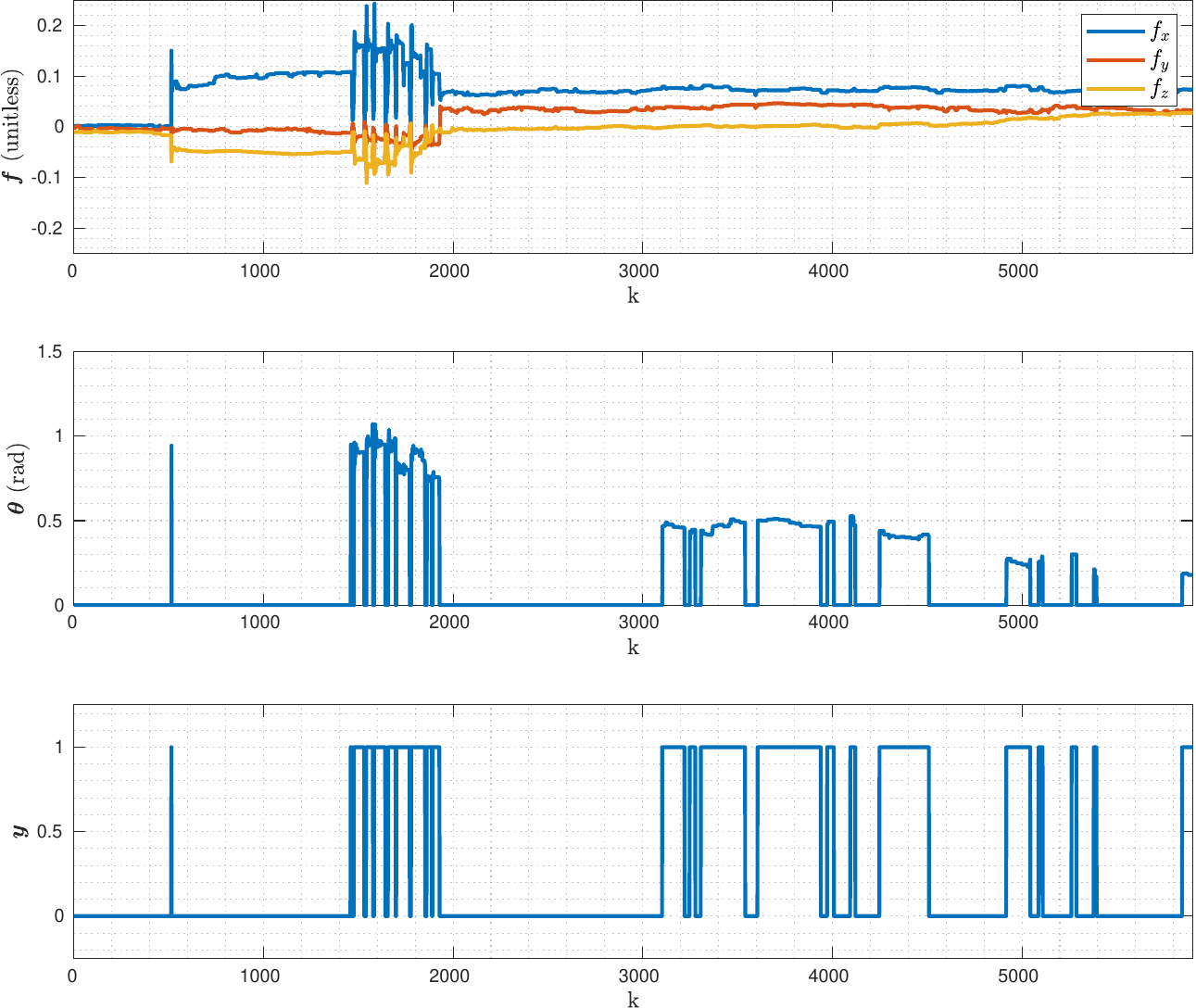}
    \caption{Results for contact force control for the Thumb}
    \label{fig:md_correction_2}
\end{figure}

Fig.~\ref{fig:id_correction_1} to Fig.\ref{fig:md_correction_2} show the results of the contact force control for the Index and Thumb fingers. Each figure comprises three plots, each a function of the time step $k$. The sampling frequency is $150 [Hz]$. The plots from top to bottom respectively track the evolution of the three components of $f$ (Eq.~\ref{eq_force}), the error angle $\theta$ from the axis angle representation of Eq.~\ref{eq:delta0} and the contact state $y$ defined in Eq.~\ref{eq_contact}.
Fig.~\ref{fig:id_correction_1} shows the result for the Index finger. We observe that during stable contact intervals ($y=1$), the angle error $\theta$ undergoes initial exponential decay, stabilizing at 0.2 radians without further correction.

A comparable structure can be observed for the Thumb in Fig. \ref{fig:md_correction_1}. $\theta$ decreases during stable contact, but the decay is less smooth than the Index finger case. The figure further illustrates a loss of contact, which is subsequently regained, allowing $\theta$ to continue decreasing until it reaches 0. 

Fig. \ref{fig:md_correction_2} provides additional insights into the Thumb's contact force control. Notably, different behavior is observed with a more frequent occurrence of contact loss. Despite this, we can observe an overall decrease in the angle error $\theta$.

We notice that gains $K_s$ and $K_\theta$ in our control system depend highly on the contact location. Notably, changes in the contact point necessitate a recalibration of these gains. In our experimentation, certain configurations allowed us to fine-tune these parameters, yielding a desired control behavior for force direction. 
Not only are the gains highly dependent on the contact location, but they also exhibit a strong interdependence. Achieving an optimal balance is challenging, as $K_s$ must be sufficiently high to maintain the contact point yet low enough to facilitate rotation around it. Similarly, $K_\theta$ should be high enough to allow for rotation around the contact point and low enough to prevent contact loss. This intricate relationship complicates the determination of appropriate gains.

The ability to control the direction of the force is further influenced by hand kinematics. While fingers can rotate around the fingertip frame along two axes, executing two rotations around the contact point depends on the contact point's location. moreover,
Inconsistencies between the physical model and the URDF model introduce errors; these errors might be negligible in cases where precise control is not required; however, in our case, those errors may potentially cause the failure of precise torque application to achieve the desired orientation.

The fact that the tactile sensor is uncalibrated for force and the absence of a contact model introduces additional challenges. 
The pseudo-force we calculate combines the taxel measurements and their noises. The taxel measurements also carry information about the sensor's deformation during contact. Thus, filtering the measurement would lead to a loss of valuable information about the force. Therefore, the lack of calibration prevents effective noise filtering.
Furthermore, the absence of a contact model makes it difficult to relate the contact force to joint torque and obtain a contact Jacobian. However, by implementing a contact model and calibrating the sensors, we can formulate the control of the direction of the force as an optimization problem \cite{pfanne2020impedance}. This would allow us to search for a solution that aligns with the hand and contact kinematics.

\section{CONCLUSIONS }

In this work, we proposed and evaluated a method to control the direction of contact force measured by tactile sensing for individual fingers. Our approach involves gathering tactile data for each finger, estimating the contact location and force, calculating the difference between the current and desired force, and then adjusting finger movements to achieve the desired force direction.

We implemented and tested our method using Xela tactile sensors with the Allegro hand, revealing promising contact force direction control results. However, our approach faces several limitations, including the lack of sensor calibration, the absence of contact modeling, and incoherence between the physical and mathematical models used for dynamics and kinematics calculations.

To improve our work, we suggest establishing a contact model through machine learning techniques and utilizing a force sensor, such as the ATI-Nano, as a ground truth. Addressing errors between the physical and mathematical models could be achieved by employing controllers capable of handling imprecision, like compliant control or adaptive control. However, in the case of adaptive control, finding an appropriate adaptation law that considers no contact constraints and corrects contact force remains challenging. 
In conclusion, controlling contact force for individual fingers remains an open and challenging research topic. Our method highlights the existing challenges in this area, paving the way for further exploration and improvement in contact force control.

\section*{APPENDIX}



\textbf{Rodriguez Formula:}
\begin{equation}
 \mathbf{R}_\theta= I^{3\times3} +s_\theta S(w) + (1- c_\theta) S(w)^2
\end{equation}
With:
\begin{align}
    c_\theta &= \frac{\prescript{\{W\}}{}{f} . \prescript{\{W\}}{}{f_d}}{|\prescript{\{W\}}{}{f}||\prescript{\{W\}}{}{f_d}|} \\
    s_\theta &=  \frac{|\prescript{\{W\}}{}{f} \times \prescript{\{W\}}{}{f_d}|}{|\prescript{\{W\}}{}{f}||\prescript{\{W\}}{}{f_d}|} \\
    w &=\frac{ \prescript{\{W\}}{}{f} \times \prescript{\{W\}}{}{f_d}} { s_\theta }
\end{align}
$S$ is the skew symmetric matrix of $w= [w_x,w_y,w_z]^T$.\\

\textbf{End-effector Jacobian} \label{an:jaco}\\
The joint velocities $\dot{q}$ and the end-effector operational space velocity $\left[ v^T \omega^T \right] $ in the inertial frame $\{W\}$  are related as follow:
\begin{equation}
   \begin{bmatrix}
     v \\
     \omega
   \end{bmatrix}
   = \begin{bmatrix}
     J_v \\
     J_\omega \end{bmatrix}  \dot{q} 
\end{equation}
$J_v$ maps $\dot{q}$ to the linear velocities $v$, whereas $J_\omega$ maps the $\dot{q}$ to the angular velocities $\omega$. For more details on computing the analytical Jacobian, readers may refer to \cite{prattichizzo2016jacobian}.\\

\bibliographystyle{IEEEtran}
\bibliography{IEEEabrv,references}

\end{document}